\documentclass[times,twocolumn,final]{elsarticle}
\usepackage[utf8]{inputenc}
\usepackage{prletters}
\usepackage{framed,multirow}

\usepackage{amssymb}
\usepackage{graphicx}
\usepackage{color}
\usepackage{siunitx}
\usepackage{bbding}
\usepackage{booktabs}
\usepackage{array, caption, threeparttable}  
\usepackage{caption}
\usepackage{url}
\usepackage{xcolor}
\definecolor{newcolor}{rgb}{.8,.349,.1}
\journal{Pattern Recognition Letters}

\begin{document}

\title{Rethinking the task-driven feature misaligned problem in one-stage object detection}
\author[1]{Shuaizheng Hao}
\ead{uj6wa67j@wycdk.com}
\author[1]{Hongzhe Liu\corref{cor1}}
\ead{liuhongzhe@buu.edu.cn}
\author[1]{Ningwei Wang}
\author[1]{Cheng Xu}
\cortext[cor1]{corresponding author;}

\address[1]{Beijing Key Laboratory of Information Service Engineering, Beijing Union University, China}

\date{April 2022}

\begin{abstract}
Common one-stage object detectors consist of two sub-tasks: object classification and box localization, using two individual branches in head networks. The feature misalignment problem, caused by the different feature sensibilities between the two branches, hurts models' performance. In this letter, we rethink and investigate the problem in one-stage detectors and find the problem is further composed of scale-level misalignment and spatial-location misalignment specifically. Based on that observation, we propose Receptive Filed Adaptor (RFA), a simple and plug-in module for models' two branches, to augment each task's adaptability to different scale information. Further, we design a novel dynamic label assignment scheme, namely Aligned Points Sampler (APS), to dynamically mine the most spatially aligned feature points during the training procedure. The extensive experiments show that with a light cost, our proposal can consistently boost models' performance around 3 AP on MS COCO across different backbones. Our code is available at https://github.com/HaoGood/MOD.
\end{abstract}


\maketitle
\section{Introduction}
\label{sec:introduction}

Object detection is a wildly-studied field in the era of deep learning, and many modern dense-head models \cite{tian2019fcos, zhang2020bridging, lin2017focal, zhu2019feature} achieve decent performance while maintaining the simplicity of network designs. Two sub-tasks, i.e., classification and localization, are often considered among them. The former aims at searching for the discriminative features across multi-classes, while the latter's goal is to localize the object with an accurate bounding box. The difference between task drivers causes the divergence of feature sensibility. That leads to the feature misalignment problem and  the performance gap between the two tasks within the same feature points. Furthermore, the NMS-based prediction method in the dense-head detectors assumes the points with high classification confidence have high localization qualities, aggravating the problem.

There are differences lying in one-stage detectors and two-stage detectors \cite{ren2015faster, girshick2015fast} in terms of the misalignment problem. For two-stage detectors, the two fully-connected structures in the second stages help extract task-aware space-independent features. Although given a large number of iterations, multi-stage models \cite{cai2018cascade} refine better-aligned crops and alleviate the misalignment problem gradually, the cost of a heavy network is not negligible. TSD \cite{song2020revisiting} resolves the problem with a lighter cost, by extracting features from two task-aware proposals for the two tasks respectively. However, its method could not apply to modern one-stage detectors. Thus, we pose a question: is there a light-cost method to solve the misalignment problem, especially applicable in one-stage models? 

For this reason, we review the fundamental structure of the misalignment problem based on FCOS \cite{tian2019fcos}, a one-stage detector. Except for spatial-location feature misalignment as discussed in previous works \cite{song2020revisiting, wu2020iou, zhang2021varifocalnet, li2020generalized, li2021generalized}, we find there is another one, which we call scale-level feature misalignment, also leading to the phenomenon. 
To investigate the disparity of scale sensibilities in the two tasks, we adopt each scale level's average loss of the two tasks separately and jointly as the scale assignment criteria. The scale level with the least average loss is assigned for each instance. The experiments are conducted on COCO and the results are in Table ~\ref{tab:scale_misalignment}. When assigned by the average classification loss or localization loss alone, models underperform a lot less than the model assigned by the average sum of them. The disparity of models' performance signifies there is a myriad of instances whose most acute classification and localization features are not shared by the same scale levels. In other words, there is a performance gap between the two tasks on the features from the same scale levels.

\begin{table}
\small
\centering
\setlength{\abovecaptionskip}{0pt}  
\setlength{\belowcaptionskip}{10pt}
\setlength{\tabcolsep}{5.4mm}
\captionsetup{width=24em} 
\caption{Models’ performance under different scale assignment standards. “Comb-loss” denotes the combined loss of two tasks.}
\label{tab:scale_misalignment}
\begin{tabular}{cccc}
\toprule
Standard & AP & AP$_{50}$ & AP$_{75}$ \\
\cmidrule(r){1-1}
\cmidrule(r){2-4}
Cls-loss & 36.7 & 56.9 & 39.1 \\
Loc-loss & 37.2 & 55.8 & 40.1 \\
Comb-loss & 38.6 & 57.5 & 41.6 \\ 
\bottomrule
\vspace{-10pt}
\end{tabular}
\end{table}

\begin{figure}[t]	
\centering
\includegraphics[width=3in]{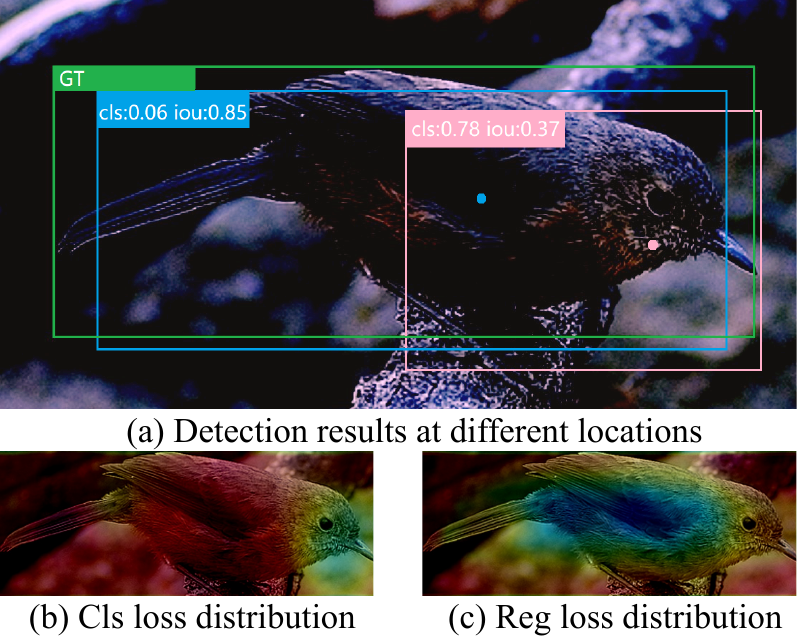}
\vspace{-3mm}
\caption{An illustration of spatial misalignment problem. Locations with red color have larger loss than locations with blue color.}
\vspace{-2mm}
\label{fig:spatial_misalignment}
\end{figure}

For the spatial-location misalignment part, we find investigating the problem from the perspective of loss distributions is instructive. Spatial distributions of the two tasks' losses within the same instance are visualized. The locations with lesser task-aware loss have better features for the task to exploit. As the example instance shown in Fig.\ref{fig:spatial_misalignment}, the two distributions are highly misaligned. The bird's head regions have the most distinctive features for the classification task, as its semantic information is the richest, while the bird's center regions are better exploited in the localization task than any other regions. These misaligned loss distributions of the two tasks show they don’t prefer the features at the same locations.

Based on the observation above, we decide to resolve the problem progressively. To alleviate the negative effect brought by the scale-level misalignment one, 
a simple and plug-in Receptive Filed Adaptor (RFA) for two individual branches is proposed. The module uses deformable convolutions to augment each task's adaption ability on scale information. 
Further, we design a novel label assignment scheme to dynamically assign the most spatially-aligned feature points with positive labels. The algorithm is called Aligned Points Sampler (APS) and is based on the analyses of spatial loss distributions. The APS helps models to predicate the reliable localization points with high classification scores, while not compromising models' classification performance. 

Our proposed model, namely Misalignment-aware One-stage Detector (MOD), is composed of the above improvements and its extra cost is minor. Extensive experiments demonstrate the sub-problems could be addressed respectively and effectively. The MOD achieves a competitive performance of 49.9 AP on COCO \textit{test-dev} in single-model and single-test configuration, with ResNeXt-101-DCN backbone.

\section{Related works}
\label{sec:related_work}

From the perspective of a general view, the methods enriching scale information for each scale level help refine task-aligned features. 
Since SSD \cite{liu2016ssd} and FPN \cite{lin2017feature} proposed to detect objects with different sizes at different scale levels, there are numerous methods to extract better-aligned features for scale levels. PaNet \cite{liu2018path}, BiFpn \cite{tan2020efficientdet}, and ASFF \cite{liu2019learning} merge scale information in a deeper and more complex manner, the performance gap between the two tasks is mitigated as each scale level includes more comprehensive information from other scale levels. Recursive-Fpn \cite{qiao2021detectors} uses ASPP \cite{chen2017deeplab} structure to merge information across all scale levels. Nas-Fpn \cite{ghiasi2019fpn} searches the best merging route horizontally and vertically from a huge search space. All these methods refine better-aligned features on scale levels, but their calculation cost is also huge.

R-CNN models \cite{ren2015faster, girshick2015fast} use fully-connected structures to extract space-independent features from the same proposals in the second stages. Compared to convolution operations restricted by spatial receptive field, fully-connected structures are not sensitive to features' locations, thus alleviating the spatial misalignment problem to some degree. Cascade R-CNN \cite{cai2018cascade} iterates the process multiple times, further refining more spatially aligned features gradually. TSD \cite{song2020revisiting} designs two task-aware proposals from the initial proposal, to decouple the spatial tangle. These R-CNN-based methods cannot directly apply to one-stage detectors. Localization Qualities Estimation (LQE) is also a common-used method. The multiplication result of the classification score and localization quality is a better standard in the NMS procedure compared to the classification score alone. Iou-Net \cite{jiang2018acquisition} is the first network learning to predict regression qualities for proposals. Following a similar paradigm, Fitness NMS \cite{tychsen2018improving} and MS R-CNN \cite{huang2019mask} learn localization qualities in two-stage methods. Iou-Aware \cite{wu2020iou} and PAA \cite{kim2020probabilistic} introduce Iou prediction branch to suggest regression-reliable points in one-stage models. Similarly, FCOS \cite{tian2019fcos} and ATSS \cite{zhang2020bridging} has a \textit{centerness} score prediction branch to represent localization qualities. Afterward, VFNet \cite{zhang2021varifocalnet} and GFL \cite{li2020generalized} merge the regression quality prediction branch directly with the classification head. 

Recently, revising the label assignment scheme is found efficient to improve performance without inference cost. ATSS \cite{zhang2020bridging} calculates Iou threshold for each instance based on the statistics of Iou between ground-truth box and its anchor boxes. PAA \cite{kim2020probabilistic} considers the sum of loss from the two tasks as the clustering standard, mining balanced sampling points and anchors in the training procedure. OTA \cite{ge2021ota} assigns labels in an overall way for each input image, its assignment results are based on optimal transportation theory. These assignment methods may inadvertently alleviate the problem, but their strategies confine positive samples to instances’ center regions.

\section{Delving into task-aligned features}
\label{sec:methods}
Our proposed method is composed of two parts. Receptive Field Adaptor (RFA) module and Aligned Points Sampler (APS) algorithm are introduced to address the scale-level and spatial-location misalignment problems accordingly. The first subsection discusses the scale-level misalignment from the view of models' receptive fields and presents the design of the RFA module in detail. In the second subsection, we elaborate on the technical details of the APS algorithm about how to mine the aligned feature points during the training procedure. 

\subsection{Receptive field adaptor}
In the heads of modern one-stage detectors, there are four sequential convolutional operations for each task-aware branch. To get feature maps with identical sizes across the two branches, each step of the four convolutional operations from the two branches shares the same convolution configuration, i.e. kernel size, stride, and padding. The receptive field of the last feature map as with each task is calculated by:
\begin{eqnarray}
\left\{ {\begin{array}{*{20}{c}}
{{\cal R}{_{cls}^{l}} = f({{\cal R}_l})}\\
{{{\cal R}_{reg}^{l}} = f({{\cal R}_l})}
\end{array}} \right.
\end{eqnarray}

${{\cal R}_l}$ is the receptive field of the initial feature map fed by the ${{l}^{th}}$ FPN level over the input image, $f( \bullet )$ is a static calculation method about receptive field across the four sequential convolution layers. Due to the same configurations in the two branches and the static $f( \bullet )$, ${\cal R}{_{cls}^{l}}$ and ${\cal R}_{reg}^{l}$ are identical and also static. As we know in the introduction section, the two tasks require different scale information within many instances. Thus the identical receptive fields compromise models' performance. 

\begin{figure}[t]	
\centering
\includegraphics[width=3in]{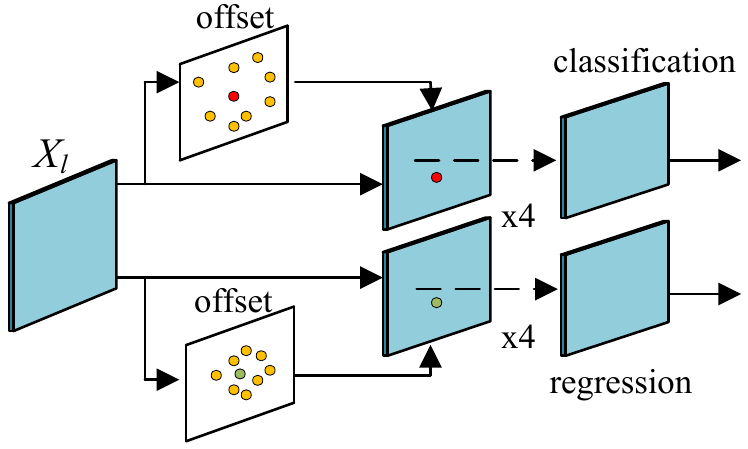}
\vspace{-3mm}
\caption{An overview of the RFA module. Two individual deformable convolutions replace the first normal convolutions in each branch.}
\vspace{-1.5mm}
\label{fig:rfa}
\end{figure}

To drive each branch to take different scale information automatically under the same scale level, as shown in Fig.\ref{fig:rfa}, we replace the first convolution operation with deformable convolution, considering the latter can dynamically adjust the receptive field. With this change, the ${\cal R}{_{cls}^{l}}$ and ${\cal R}_{reg}^{l}$ are calculated as follows:
\begin{eqnarray}
\left\{ {\begin{array}{*{20}{c}}
{{\cal R}{_{cls}^{l}} = {\cal F}({{\cal R}_l},{\theta _{cls}|{X_l}})}\\
{{{\cal R}_{reg}^{l}} = {\cal F}({{\cal R}_l},{\theta _{reg}|{X_l}})}
\end{array}} \right.    
\end{eqnarray}

${X_l}$ is the initial feature map information shared by both tasks, while ${\theta _{cls}}$ and ${\theta _{reg}}$ are learnable parameters provided by the deformable convolutions from each branch. ${\cal F}(\bullet)$ is a dynamic receptive field calculation method according to ${\cal R}_l$ and ${\theta}$. Although a simple replacement, this adjustment gives models the ability to adapt individual receptive fields for the individual tasks. Under this configuration, each task-driven branch dynamically extracts its own scale information by tweaking its receptive field according to detailed feature information and task requirement. In other words, the output features of the two tasks are better aligned on scale levels.

There are also other ways to adjust each branch’s receptive field, such as placing dilated convolution or adding extra convolutions according to detailed preferences analyses. Compared to them, our method presents a better accommodation ability. As LCDC ~\cite{mac2019learning} shows, with deformable convolution, each point of feature map has a varied receptive field in theory.



\subsection{Aligned points sampler}
In traditional label assignment algorithms, anchor-based methods use pre-defined Iou thresholds, and anchor-free models use scale-range and spatial-radius hyperparameters. These hyperparameters imply the models could not adapt to diverse instances' features automatically. In contrast, we design a single-hyperparameter label assignment scheme to assign the most aligned points with positive labels. The algorithm is driven by the losses fed back from the model itself. We will talk about it in two sequential subprocedures for a clear accountant.

\begin{figure*}[t]
\centering
\includegraphics[width=7in]{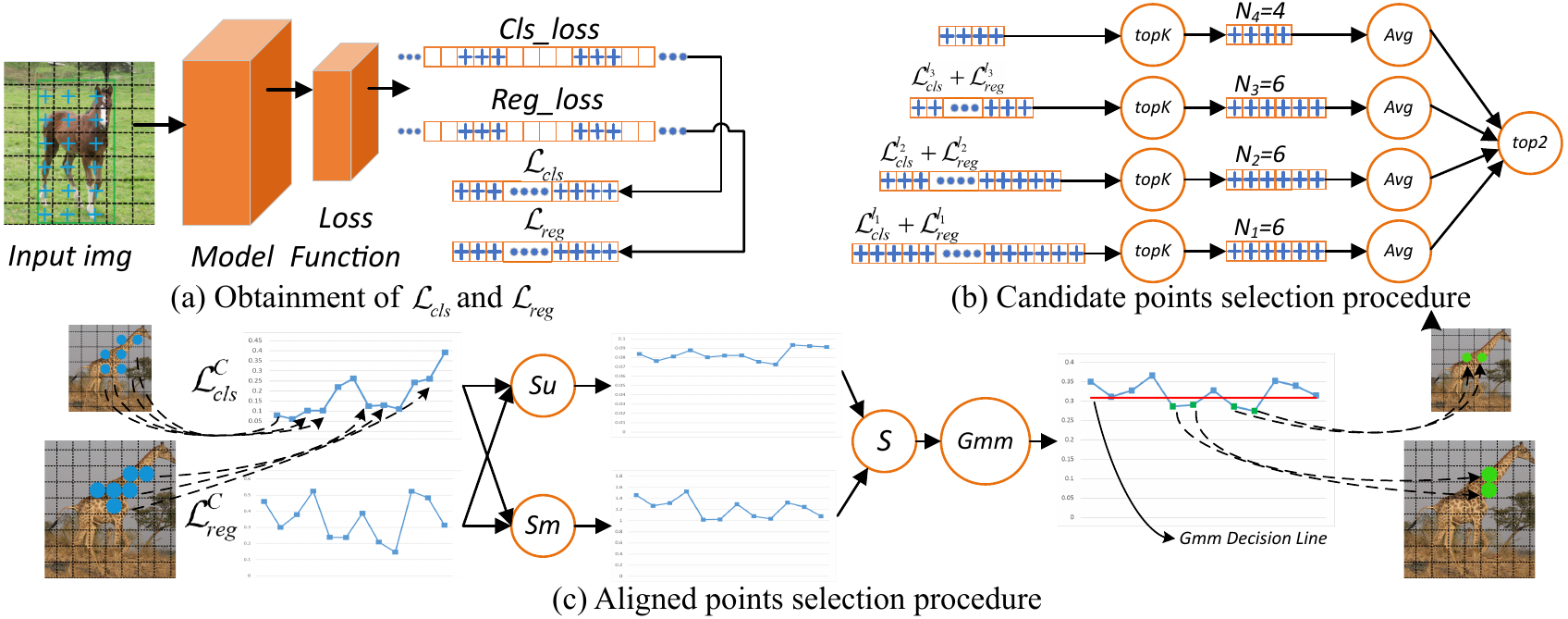}
\vspace{-3mm}
\caption{An overall view of the sampling process of the APS. In (a) blue crosses denote the initially positive points ${\cal P}$. Instead of summing up losses of all points, we select out the losses of positive points ${\cal P}$. In (b) each row represents the positive loss vector corresponding to each scale level. The topK selects the points within each level, the top2 selects the scale levels with the smallest average loss. 
In (c) blue circles are final candidate points ${C_{{l^*}}}$ across scale levels, green circles are final positive points, the vectors ${\cal L}_{cls}^{\cal C}$, ${\cal L}_{reg}^{\cal C}$, .etc are presented in the real detailed values. ${\cal L}_{cls}^{\cal C}$, ${\cal L}_{reg}^{\cal C}$ are converted to ${S_u}$ and ${S_m}$, which are further calculated into $S$. Then the GMM algorithm uses $S$ to output positive points cluster.}
\vspace{-1mm}
\label{fig:aps}
\end{figure*}

\subsubsection{Candidate points selection procedure}
For each instance ${\cal I}$ from the input image, we define ${{\cal P}_l}$ as the feature points spatially within the bounding box of ${\cal I}$ in the ${{l}^{th}}$ scale level. Initially, ${{\cal P}_l}$ are all labeled as positive as shown in Fig.\ref{fig:visual_aps}.(a),
and we get the loss vectors ${\cal L}_{cls}$ and ${\cal L}_{reg}$ fed back from the forward propagation. Note that ${\cal L}_{cls}$ and ${\cal L}_{reg}$ are loss vectors corresponding to the initially positive points ${\cal P}$. And they are not the final losses for backpropagation, only used for label assignment.

The candidate points are could-be positive samples. The first goal is to determine the number of candidate points ${N_l}$ of the ${{l}^{th}}$ scale level. ${N_l}$ is circumscribed by the total available candidate points and a maximum hyperparameter ${\cal K}$:
\begin{eqnarray}
{N_l} = \max ({\cal K},\sum i |P_i \in {\cal P}_l)
\end{eqnarray}

In modern assignment schemes, positive candidate points are confined to instances' center regions. This hurts models' performance as aligned points do not necessarily lie within center regions. In contrast, we select the initial candidate points ${\cal C}_l$ for the ${{l}^{th}}$ scale level simply according to the combined loss of two tasks, without spatial confinement:
\begin{eqnarray}
	{\cal C}_l = \{ \mathop {\arg \min }\limits_{P_i \in {\cal P}_l,\left| {P_i} \right| = {N_l}} \sum\limits_{i} \left( {{\cal L}_{cls}^i + {\cal L}_{reg}^i} \right)\}
\end{eqnarray}

The ${\cal L}_{cls}^{i}$ and ${\cal L}_{reg}^{i}$ are loss elements of the specific levels and points. Basically, the above formula denotes a topK selection procedure. The condition "${|P_{i}|}={N}_{l}$" limits the ${N}_{l}$ feature points with the smallest losses are chosen as candidate points. 

Given ${\cal C}_l$, the next goal is to determine candidate levels $l^*$ from all scale levels $L$. We assign two scale levels for each instance to meet the demand of model’s generality on the scale dimension. In our case, instead of Iou thresholds or scale range hyperparameters, we assign directly by the average positive loss of each scale level forwarded by the ${\cal C}_l$. The scale assignment results ${l^*}$ are chosen by: 
\begin{eqnarray}
	l^* = \{\mathop {\arg \min }\limits_{l \in L,\left| l \right| = 2} \sum\limits_{{C_i} \in {C_l}} \sum\limits_{P_{j} \in {C_i}} {({\cal L}_{cls}^j + {\cal L}_{reg}^j)/{N_l}} \}
\end{eqnarray}

In the above formula, the two scale levels with the smallest positive average losses are chosen. That is given by the intuition: given one instance, the average positive loss reflects the preference degree for the model itself about each scale level. Under this assignment standard, the assigned scale levels are no longer purely driven by instances’ sizes but determined by models’ preferences automatically. At last, the final candidate points ${C_{{l^*}}}$ for the next procedure is determined.

\subsubsection{Aligned points selection procedure}
Given the final candidate points ${C_{{l^*}}}$ for each instance ${\cal I}$, the task is to further mine the most spatially aligned points from them. Let ${\cal L}_{cls}^{\cal C}$ and ${\cal L}_{reg}^{\cal C}$ be the loss vectors corresponding to ${C_{{l^*}}}$. There are two metrics to be considered for each candidate point: (1) the overall unfitness degree ${S_u}$ under consideration of both tasks; (2) the misaligned degree ${S_m}$ caused by spatially misaligned loss distributions as we mentioned in the introduction section. As ${\cal L}_{cls}^{\cal C}$ and ${\cal L}_{reg}^{\cal C}$ reflect model's preference degree for respective tasks naturally, ${S_u}$ and ${S_m}$ are evaluated by pair-wise loss vectors (${{\cal L}_{cls}^{\cal C}}$,${{\cal L}_{reg}^{\cal C}}$) as follows:
\begin{equation}
{S_u} = ( {\frac{{\exp ({{\cal L}_{cls}^{\cal C}})}}{{\sum\nolimits_k {\exp ({\cal L}_{cls}^{{\cal C},k})}}} + \frac{{\exp ({{\cal L}_{reg}^{\cal C}})}}{{\sum\nolimits_k {\exp ({\cal L}_{reg}^{{\cal C},k})} }}} )/2
\end{equation}
\begin{equation}
{{S}_{m}}=\frac{1}{1+{{e}^{-|{{\mathcal{L}}_{cls}^{\cal C}}-{{\mathcal{L}}_{reg}^{\cal C}}|}}}
\end{equation}

For unfitness score ${S_u}$, we use the softmax function above to evaluate ${{\cal L}_{cls}^{\cal C}}$ and ${{\cal L}_{reg}^{\cal C}}$ separately and redistribute them into the same measurable standard, that is given by the advantage softmax function is monotone and the sum of its output is one. For misaligned score ${S_m}$, as shown in the Fig.\ref{fig:aps}.(c), the difference between ${{\cal L}_{cls}^{\cal C}}$ and ${{\cal L}_{reg}^{\cal C}}$ fluctuates a lot between candidate points, so we choose the above sigmoid function to convert them into a rather unified output. 

In essence, ${S_u}$ evaluates the sum of the two task-aware losses. This helps ${S_u}$ reflect the models' exploitation feasibility of the certain points' features. By assessing the difference between the losses, ${S_m}$ presents the performance gap between the two tasks. The above calculation functions in the formulas limit the values of ${S_u}$ and ${S_m}$ within the range zero to one. Note that both of them should be small for a high-quality positive point and they are also vectors of the same size as ${{\cal L}_{cls}^{\cal C}}$ and ${{\cal L}_{reg}^{\cal C}}$.

If only assigning labels according to the ${S_u}$, the model may be involved in the over-fitting problem as only the easiest samples are assigned with positive labels. On the other side, if only assigning labels according to ${S_m}$, the model would fail to converge during training as it would treat candidate points with high but close pair-wise loss as positive. Given the analyses above, we simply define the final score of candidate points ${C_{{l^*}}}$ as the square root of the multiplication of unfitness score ${S_u}$ and misalignment score ${S_m}$:
\begin{equation}
S = \sqrt {{S_u}*{S_m}}
\end{equation}

The score $S$ is both distribution-aware and misalignment-aware of the two tasks. The decision of final positive points and negative points from candidate points ${C_{{l^*}}}$ according to $S$ is based on Gaussian Mixture Model (GMM) cluster algorithm, as shown in Fig.\ref{fig:aps}.(c). The two-proponent GMM assumes the inputs are composed of two parts, which are positive samples and negative samples in our case. Compared to the topK manner, GMM has a better performance as it has the ability to dynamically cluster out positive points and negative points according to scores distribution, which is suitable for our appliance.

\section{Experiments and analyses}
\label{sec:experiments}
In this section, the ablation studies and extensive experiments are presented. We first introduce the implementation details of experiments in ~\ref{subsec:imp_details}. Then the ablation studies are introduced in ~\ref{subsec:indi_effects}-~\ref{subsec:assign_analyses}, including validating the effects of individual components, analyzing the effects of the RFA module, and anatomizing the assignment procedure of the APS algorithm. Afterward, we apply our proposal to other one-stage models in  ~\ref{subsec:trasfer_analyses} and compare it with other detection models in ~\ref{subsec:sota_comparison}. Finally, the visualization example of our model is presented in ~\ref{subsec:visual_result}.

\subsection{Implementation details}
\label{subsec:imp_details}
The training and evaluation dataset is the COCO benchmark. Models' performance in ablation studies is evaluated on val set. And when compared to other methods in \ref{subsec:sota_comparison}, models' performance is evaluated on \textit{test-dev} set.

We use ResNet-50 pre-trained on ImageNet and FPN as our backbone and neck for our ablation studies. Synchronized stochastic gradient (SGD) is employed on 8 GPUs with mini-batch set to 16. The models are trained with 90k iterations. Following other common practices, we use Focal Loss \cite{lin2017focal} and Giou Loss \cite{rezatofighi2019generalized} for the classification task and regression task respectively. As recent researches \cite{li2020generalized, li2021generalized} shows, \textit{centerness} has some shortcomings for predicting regression qualities. We simply replace the supervision signal with Giou. As for the label assignment scheme, we set ${\cal K}$ to 9 and GMM’s minimum and maximum scores of the candidate points are set by the means of the two components.

\subsection{Effects of individual components}
\label{subsec:indi_effects}
In this part, we verify the effectiveness of each component of our proposal. Receptive field Adaptor (RFA) and Aligned Point Sampler (APS) are gradually added to the baseline to make comprehensive comparisons and analyses. As shown in the second row in Table 2, without complex network designs and a bunch of extra cost, only RFA brings a significant performance improvement around 2 AP, which suggests the scale misalignment is alleviated by a huge progress. The third row of Table~\ref{tab:abl_all} shows the APS also lifts the performance by 1.8 AP in a inference cost-free way. The combined result of RFA and APS gets an absolute improvement of 3.1AP, showing both the simplicity and effectiveness of our proposal MOD. Note that the FPS (Frames Per Second) of our proposed method during inference is only reduced by 0.2 compared to baseline, suggesting the calculation cost of the extra deformable convolution in the RFA module is minor and the proposed method is efficient.

\subsection{Effect of RFA reducing scale misalignment phenomenon}
As we talked about in the introduction section, there is a performance gap between the two tasks on the features from the same scale levels. 
Note that each task's average loss within the same scale levels reflects the models' preference degree directly. Thus we use the loss gap (i.e. difference between the two losses) and loss sum of the two tasks from the assigned scale levels to represent models’ adaption ability and overall performance on scale information respectively. 
They are collected at the model's different training iterations. 

The results are shown in Table~\ref{tab:loss_gap} and Table~\ref{tab:loss_sum}. Models with RFA module consistently get around $2{\rm{\% }}$ relative reduction on the two metrics at different training iterations compared to baseline models. That suggests RFA’s consistent ability on conciliating scale misalignment between the two tasks as the loss gaps are mitigated obviously, and this property does not compromise the models' overall performance as the loss sums of two tasks are also reduced.

\begin{table}[t]
\small
\centering
\setlength{\abovecaptionskip}{0pt}  
\setlength{\belowcaptionskip}{10pt}
\setlength{\tabcolsep}{3.1mm}
\caption{Ablation results of Receptive field Adaptor (RFA) and Aligned Points Sampler (APS). “*” denotes \textit{centerness} as auxiliary branch, while others use Giou prediction.} \label{tab:abl_all}
\begin{tabular}{lcccccc}
	\toprule
	Method &RFA &APS &FPS &AP &AP$_{50}$ &AP$_{75}$
	\\
	\cmidrule(r){1-1}
	\cmidrule(r){2-3}
	\cmidrule(r){4-4}
	\cmidrule(r){5-7}
	FCOS* & & & 14.7 & 38.6 & 57.2 & 41.9 \\
	FCOS  & & & 14.7 & 39.4 & 57.3 & 42.7 \\
	\cmidrule(r){1-1}
	\cmidrule(r){2-3}
	\cmidrule(r){4-4}
	\cmidrule(r){5-7}
	&\Checkmark & & 14.5 & 40.5 & 58.0 & 44.3 \\
	MOD&  &\Checkmark & 14.7 & 40.4 & 58.2 & 43.9 \\
		   &\Checkmark &\Checkmark & 14.5 & \textbf{41.7} & \textbf{59.3} & \textbf{45.1} \\
	\bottomrule
\end{tabular}
\vspace{-4mm}
\end{table}

\begin{table}[!t]
\small
\centering
\setlength{\abovecaptionskip}{0pt}  
\setlength{\belowcaptionskip}{10pt}
\setlength{\tabcolsep}{3.1mm}
\caption{Statistics of loss gap between the two tasks at model's different iterations} \label{tab:loss_gap}
\begin{tabular}{lccccc}
	\toprule
	Iterations &15k &30k &45k &60k &75k	\\
	\cmidrule(r){1-1}
	\cmidrule(r){2-6}
	FCOS &0.169  &0.156 &0.141 &0.150 &0.137	\\
    FCOS+RFA &\textbf{0.167} &\textbf{0.153} &\textbf{0.139}  &\textbf{0.146} &\textbf{0.134}	\\
	Reduction&	1.2${\rm{\% }}$& 1.9${\rm{\% }}$ &1.4${\rm{\% }}$ &2.7${\rm{\% }}$ &2.2${\rm{\% }}$	\\
	\bottomrule
\end{tabular}
\vspace{-4mm}
\end{table}

\begin{table}[!t]
\small
\centering
\setlength{\abovecaptionskip}{0pt}  
\setlength{\belowcaptionskip}{10pt}
\setlength{\tabcolsep}{3.1mm}
\caption{Statistics of loss sum between the two tasks at model's different iterations.} \label{tab:loss_sum}
\begin{tabular}{lccccc}
	\toprule
	Iterations &15k &30k &45k &60k &75k	\\
	\cmidrule(r){1-1}
	\cmidrule(r){2-6}
	FCOS &0.580 &0.523 &0.487 &0.454 &0.412\\
	FCOS+RFA &\textbf{0.570} &\textbf{0.511} &\textbf{0.466} &\textbf{0.442} &\textbf{0.401}	\\
	Reduction &1.2${\rm{\% }}$ &2.3${\rm{\% }}$ &4.5${\rm{\% }}$ &2.5${\rm{\% }}$ &2.7${\rm{\% }}$	\\
	\bottomrule
\end{tabular}
\vspace{-4mm}
\end{table}

\subsection{Effects of scale and spatial assignment in APS}
\label{subsec:assign_analyses}
The APS is composed of two subprocedures. Scale levels are assigned in the former and spatial points are chosen in the latter. To verify the effects of these two assignment procedures, we implement them individually and jointly. The experiments are conducted on models with the RFA module to make a clear comparison. As Table~\ref{tab:abl_aps} shows, when the two procedures are separately adopted, the improvement of models' performance is minor with 0.3 AP and 0.5 AP boost respectively. However, when the two procedures are combined, the model reaches 41.7 AP with an absolute 1.2 AP gain. This synergic effect means each procedure helps the other find the best-aligned results on corresponding scale levels and spatial locations respectively.

\subsection{Transferred effects of proposal}
\label{subsec:trasfer_analyses}
The RFA module are easily plug-in to other one-stage detectors. The APS algorithm assigns one sample point with one label, so it is naturally applicable in anchor-free models. However, for anchor-based detectors, caused by the three anchors per feature point, there are some adjustments for implementing the APS algorithm. We simply assign each feature point with only one anchor, by choosing the one with the smallest loss out of the three. The results of our MOD proposal applied on other one-stage detectors are shown in Table~\ref{tab:cmp_sota}. With a minor cost, the MOD also lifts other one-stage detectors by around 3 AP, suggesting the generality of our proposal.

\begin{table}[t]
\small
\centering
\setlength{\abovecaptionskip}{0pt}  
\setlength{\belowcaptionskip}{10pt}
\setlength{\tabcolsep}{2.6mm}
\captionsetup{width=24.65em} 
\caption{Ablation results of scale assignment and spatial assignment procedures.} \label{tab:abl_aps}
\begin{tabular}{ccccc}
	\toprule
	Scale-level &Spatial-location &AP &AP$_{50}$ &AP$_{75}$	\\
	\cmidrule(r){1-2}
	\cmidrule(r){3-5}
	&	&40.5   &58.0   &44.3 \\
	\Checkmark &	    &40.8   &58.9   &44.2 \\
	&\Checkmark &41.0	&58.8   &45.0 \\
	\Checkmark &\Checkmark &\textbf{41.7} &\textbf{59.3} &\textbf{45.1}	\\
	\bottomrule
\end{tabular}
\vspace{-4mm}
\end{table}

\begin{table}[t]
\small
\centering
\setlength{\abovecaptionskip}{0pt}  
\setlength{\belowcaptionskip}{10pt}
\setlength{\tabcolsep}{3.2mm}
\captionsetup{width=24.65em} 
\caption{Transferred results on other networks.} \label{tab:cmp_sota}
\begin{tabular}{lcccc}
\toprule
Methods &FPS &AP & AP$_{50}$ & AP$_{75}$ \\
\cmidrule(r){1-1}
\cmidrule(r){2-2}
\cmidrule(r){3-5}
Retinanet \cite{lin2017focal} &15.4 &36.5	&55.4	&39.1	\\
Retinanet+MOD &15.0 &\textbf{39.8}	&\textbf{58.3}	&\textbf{42.9}	\\
\cmidrule(r){1-1}
\cmidrule(r){2-2}
\cmidrule(r){3-5}
Foveabox \cite{kong2020foveabox} &15.6 &36.5  &56.0  &38.6	\\
Foveabox+MOD &15.1 &\textbf{39.1} &\textbf{58.1}  &\textbf{42.1}	\\
\bottomrule
\end{tabular}
\vspace{-4mm}
\end{table}

\begin{table}[!t]
\small
\centering
\setlength{\abovecaptionskip}{0pt}  
\setlength{\belowcaptionskip}{10pt}
\caption{Detection results on the COCO \textit{test-dev} set.} \label{tab:7}
\begin{tabular}{lcccccc}
	\toprule
	Method &AP &AP$_{50}$ &AP$_{75}$ &AP$_{S}$ &AP$_{M}$ &AP$_{L}$	\\
	\cmidrule(r){1-1}
	\cmidrule(r){2-4}
	\cmidrule(r){5-7}
	\multicolumn{7}{c}{ResNet-101 backbone} \\
	Cascade R-CNN \cite{cai2018cascade} &42.8 &62.1 &46.3 &23.7 &45.5 &55.2 \\ 
	TSD           \cite{song2020revisiting} &43.2 &64.0 &46.9 &24.0 &46.3 &55.8 \\
	FCOS w/ imprv \cite{tian2019fcos} &41.5 &60.7 &45.0 &24.4 &44.8 &51.6 \\
	FSAF          \cite{zhu2019feature} &40.9 &61.5 &44.0 &24.0 &44.2 &51.3 \\
    Noisy Anchor  \cite{li2020learning}&41.8 &61.1 &44.9 &23.4 &44.9 &52.9 \\
	MAL           \cite{ke2020multiple} &43.6 &62.8 &47.1 &25.0 &46.9 &55.8 \\
	SAPD          \cite{zhu2020soft} &43.5 &63.6 &46.5 &43.5 &63.6 &46.5 \\
	ATSS          \cite{zhang2020bridging} &43.6 &62.1 &47.4 &26.1 &47.0 &53.6 \\
	PAA           \cite{kim2020probabilistic} &44.8 &63.3 &48.7 &26.5 &48.8 &56.3 \\
	GFL           \cite{li2020generalized} &45.0 &63.7 &48.9 &27.2 &48.8 &54.5 \\
	GFLV2         \cite{li2021generalized} &46.2 &64.3 &\textbf{50.5} &27.8 &49.9 &57.0 \\
	\textbf{MOD(ours)} &\textbf{46.3} &\textbf{64.3} &50.3 &\textbf{27.9} &\textbf{50.2} &\textbf{57.9} \\
	\cmidrule(r){1-1}
	\cmidrule(r){2-4}
	\cmidrule(r){5-7}
	\multicolumn{7}{c}{ResNeXt-101-64x4d backbone} \\
	SAPD         \cite{zhu2020soft} &45.4 &65.6 &48.9 &27.3 &48.7 &56.8 \\
	ATSS         \cite{zhang2020bridging} &45.6 &64.6 &49.7 &28.5 &48.9 &55.6 \\
	GFL          \cite{li2020generalized} &46.0 &65.1 &50.1 &28.2 &49.6 &56.0 \\
	PAA          \cite{kim2020probabilistic} &46.6 &65.6 &50.8 &28.8 &50.4 &57.9 \\
	IQDet        \cite{ma2021iqdet} &47.0 &65.7 &51.1 &29.1 &50.5 &57.9 \\
	OTA          \cite{ge2021ota} &47.0 &65.8 &51.1 &29.2 &50.4 &57.9\\
	\textbf{MOD(ours)} &\textbf{48.5} &\textbf{67.1} &\textbf{52.6} &\textbf{30.7} &\textbf{52.2} &\textbf{60.2} \\
	\cmidrule(r){1-1}
	\cmidrule(r){2-4}
	\cmidrule(r){5-7}
	\multicolumn{7}{c}{ResNeXt-101-64x4d-DCN backbone} \\
	SAPD         \cite{zhu2020soft} &47.4 &67.4 &51.1 &28.1 &50.3 &61.5 \\
	ATSS         \cite{zhang2020bridging} &47.7 &66.5 &51.9 &29.7 &50.8 &59.4 \\
	GFL          \cite{li2020generalized} &48.2 &67.4 &52.6 &29.2 &51.7 &60.2 \\
	PAA          \cite{kim2020probabilistic} &49.0 &67.8 &53.3 &30.2 &52.8 &62.2 \\
	IQDet        \cite{ma2021iqdet} &49.0 &67.5 &53.1 &30.0 &52.3 &62.0 \\
	OTA          \cite{ge2021ota} &49.2 &67.6 &53.5 &30.0 &52.5 &62.3 \\
	\textbf{MOD(ours)} &\textbf{49.9} &\textbf{68.6} &\textbf{54.2} &\textbf{31.2} &\textbf{53.6} &\textbf{62.8} \\
	\bottomrule
\end{tabular}
\vspace{-4mm}
\end{table}

\subsection{Comparison with other models}
\label{subsec:sota_comparison}
We compare our proposed MOD with other models dedicated to resolving misalignment problems on COCO \textit{test-dev}, results are in Tabel~\ref{tab:7}. All models are trained on 2x learning schedule and with MSTrain strategy. For a fair comparison, we only report results of single model and single testing scale. With backbone ResNet-101, our model achieves 46.3 AP, outperforming R-CNN based models and LQE-based models, such as GFL ~\cite{li2020generalized} (by $\sim$1.5AP) and MAL ~\cite{ke2020multiple} (by $\sim$3AP). This performance also surpasses other labels-refined models, such as ATSS ~\cite{zhang2020bridging} (by $\sim$3AP) and PAA ~\cite{kim2020probabilistic} (by $\sim$2AP). The advantage continues on other backbones. Note that the extra calculation cost of our proposal is minor, so it's remarkable to achieve such a huge improvement. These comparison results suggest the effectiveness of our proposed strategy to resolve scale-level and spatial-location misalignment problems progressively. On ResNeXt-101-64x4d-DCN backbone, our single-model and single-test method achieves a very competitive performance with 49.9 AP.

\subsection{Visualization example of APS’ spatial assignment results}
\label{subsec:visual_result}
\begin{figure}[!t]
\centering
\includegraphics[width=3in]{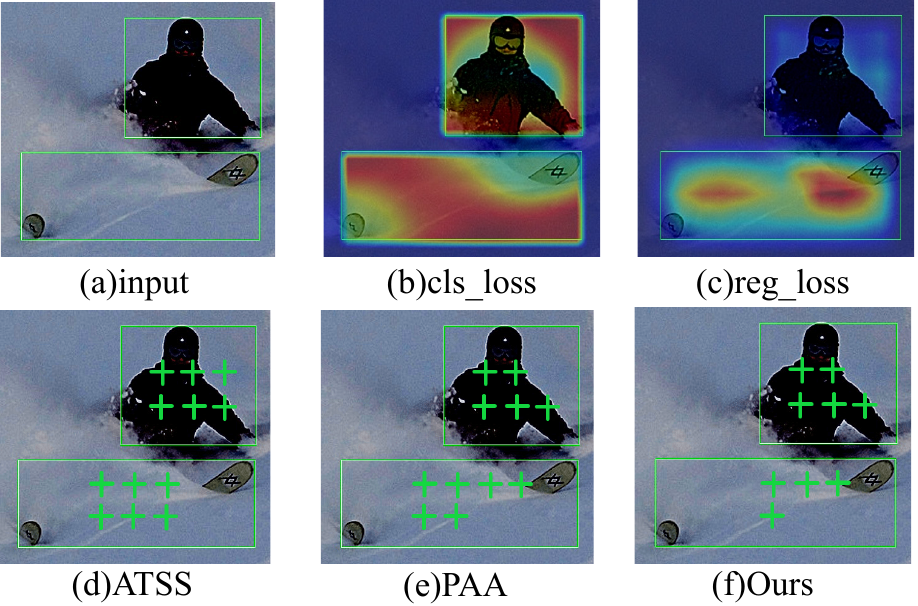}
\vspace{-3mm}
\caption{The visualization of spatial label assignment. The first row shows the input image and loss distributions of the two tasks. The green crosses in the second row are positive assignment points.}
\label{fig:visual_aps}
\end{figure}

We visualize the spatial distributions of the two tasks' losses, and further mark the positive assignment results in this experiment. We choose a representative and informative example and the visualization results are shown in Fig~\ref{fig:visual_aps}. There are two instances in the example image. According to the loss distributions of the two tasks, the bottom skating board is spatially misaligned while the top human is fairly spatially aligned. Compared to other improved label assignment strategies, such as ATSS ~\cite{zhang2020bridging} and PAA ~\cite{kim2020probabilistic}, there isn’t a significant assignment difference in the spatially aligned instance. However, in the spatially misaligned instance, our proposed APS algorithm assigns more aligned points with positive labels according to the loss distribution of each task, than ATSS ~\cite{zhang2020bridging} and PAA ~\cite{kim2020probabilistic} do. Based on the above analyses, the APS algorithm assigns more spatially aligned points as positive while training. In inference procedure, this enhances the model's ability to detect aligned feature points, whose regression quality and classification accuracy are both guaranteed. This property alleviates the negative effect brought by the task-driven spatially misaligned problem in object detection effectively. 

\section{Conclusion}
In this paper, we rethink and investigate the fundamental structure of the feature misalignment problem in one-stage detectors. We analyze the problem as composed of scale and spatial. For scale misalignment, we propose a plug-in and light module RFA to tweak each branch’s receptive field. For spatial misalignment, we propose a novel label assignment scheme mining the most spatially aligned points. Experiments show that our proposal significantly boosts models performance and the corresponding misalignment problems are alleviated.

\section{Acknowledgments}
The authors thank the financial support from NNSFC (Grant NO.61871039, 62102033, 62171042, 62006020, and 61906017).
\bibliographystyle{elsarticle-num}
\bibliography{references}

\end{document}